# Diffusion-Based Inverse Design of Dielectric Resonator Metasurfaces for Shaping Smart Electromagnetic Environments

M. Tsukerman*, K. Grotov, D. Vovchuk, and P. Ginzburg

## Abstract

Future wireless systems are expected to transform the surrounding space from a passive propagation medium into a smart electromagnetic environment, where engineered surfaces control wave propagation, support wireless sensing, and create programmable electromagnetic fingerprints. A key challenge in realizing this vision is the inverse design of metasurfaces for tailored electromagnetic propagation. While forward analysis evaluates the response of a known geometry, the inverse task starts from a prescribed scattering signature and seeks a physically realizable structure that produces it. This inverse task is inherently nonlinear and often high-dimensional, while candidate solutions may be non-unique and provide no direct indication of practical realizability. Here, we introduce a conditional diffusion framework for inverse design of dielectric resonator metasurfaces from target angular scattering patterns. Trained on T-matrix simulated geometry–response pairs, the model learns a conditional distribution of geometries instead of a deterministic mapping, enabling multiple candidate designs for the ill-posed inverse problem. The best generated metasurface achieves a mean percentage error of 1.39%, outperforming CMA-ES optimization (4.1% after 10 h) while requiring only about one minute for after-training inference. The model also produces lower error distributions than deterministic neural baselines for out-of-distribution spectra, highlighting the potential of diffusion models for efficient metasurface design.

*corresponding author (mik.tsukerman@gmail.com)

## Introduction

Future wireless systems are expected to operate in environments that are no longer treated as passive propagation media. Instead, the surrounding space can become an active electromagnetic component, shaping waves in ways that support communication, sensing, localization, and identification. This vision is captured by the concept of smart electromagnetic environments, where engineered surfaces are embedded into physical spaces to control reflections, redistribute energy, shape multipath, and create object- or task-specific electromagnetic fingerprints [1], [2], [3], [4], [5].

Metasurfaces are among the main physical platforms proposed for this purpose. By arranging subwavelength or resonant elements across a surface, they can impose tailored amplitude, phase, polarization, or scattering responses on incident waves [6], [7], [8], [9]. This capability has motivated a broad range of approaches, including passive metasurfaces with fixed responses, reconfigurable intelligent surfaces with tunable elements, and optimization-driven designs aimed at beam steering, focusing, anomalous reflection, channel improvement, or sensing enhancement [4], [5], [10]. In most cases, however, the design procedure is still formulated around a reduced target: a prescribed phase profile, a desired far-field lobe, or an isolated scattering characteristic.

A key difficulty is that smart electromagnetic environments require more than designing a standalone surface under idealized illumination. The surface response must remain physically realizable, compatible with material and fabrication constraints, and useful inside a real propagation scenario containing multiple objects, users, paths, and boundary conditions [11]. This creates a demanding inverse-design task. Forward electromagnetic analysis can evaluate the scattering response of a known geometry, but the practically relevant direction is reversed: starting from a desired electromagnetic signature or environmental function, one must identify a feasible structure that produces it. Such a task is nonlinear, often high-dimensional, and constrained by the fact that mathematically suitable responses do not automatically correspond to manufacturable or robust designs [10], [12], [13], [14], [15].

Traditional approaches, such as topology optimization or evolutionary algorithms, rely on repeated forward simulations and often require expert tuning, making them computationally demanding for rapid inverse design [16]. These challenges motivate data-driven inverse scattering methods that learn from electromagnetic simulations while preserving the ability to generate multiple candidate geometries [17]. Machine Learning has provided a promising alternative allowing for learning the structure–response mapping [18]. In particular, generative models are attractive because they do not force the inverse map into a single deterministic prediction. Instead, they approximate the data distribution information about which can be extracted from the training dataset [19]. After approximation, generative models sample a family of possible structures conditioned on a target response without versatility vanishing and falling to a mean solution that does not satisfy all conditions [20]. This capability is especially relevant for metasurface design in smart electromagnetic environments, where the final goal is not merely to reproduce an isolated scattering curve, but to enable controllable electromagnetic functionality under realistic system-level requirements [21], [22].

Here, we position diffusion-based inverse design as a generative route toward metasurface synthesis for smart electromagnetic environments. Rather than treating metasurface design as a search for one optimized structure under a fixed target, the proposed approach frames the task as conditional generation of feasible scattering architectures. This shift is important for complex electromagnetic settings, where useful solutions may form a family of possible designs rather than a single optimum. By combining data-driven generation with forward electromagnetic verification, such a workflow can bridge the gap between abstract target responses and practically testable metasurface candidates, Figure 1.

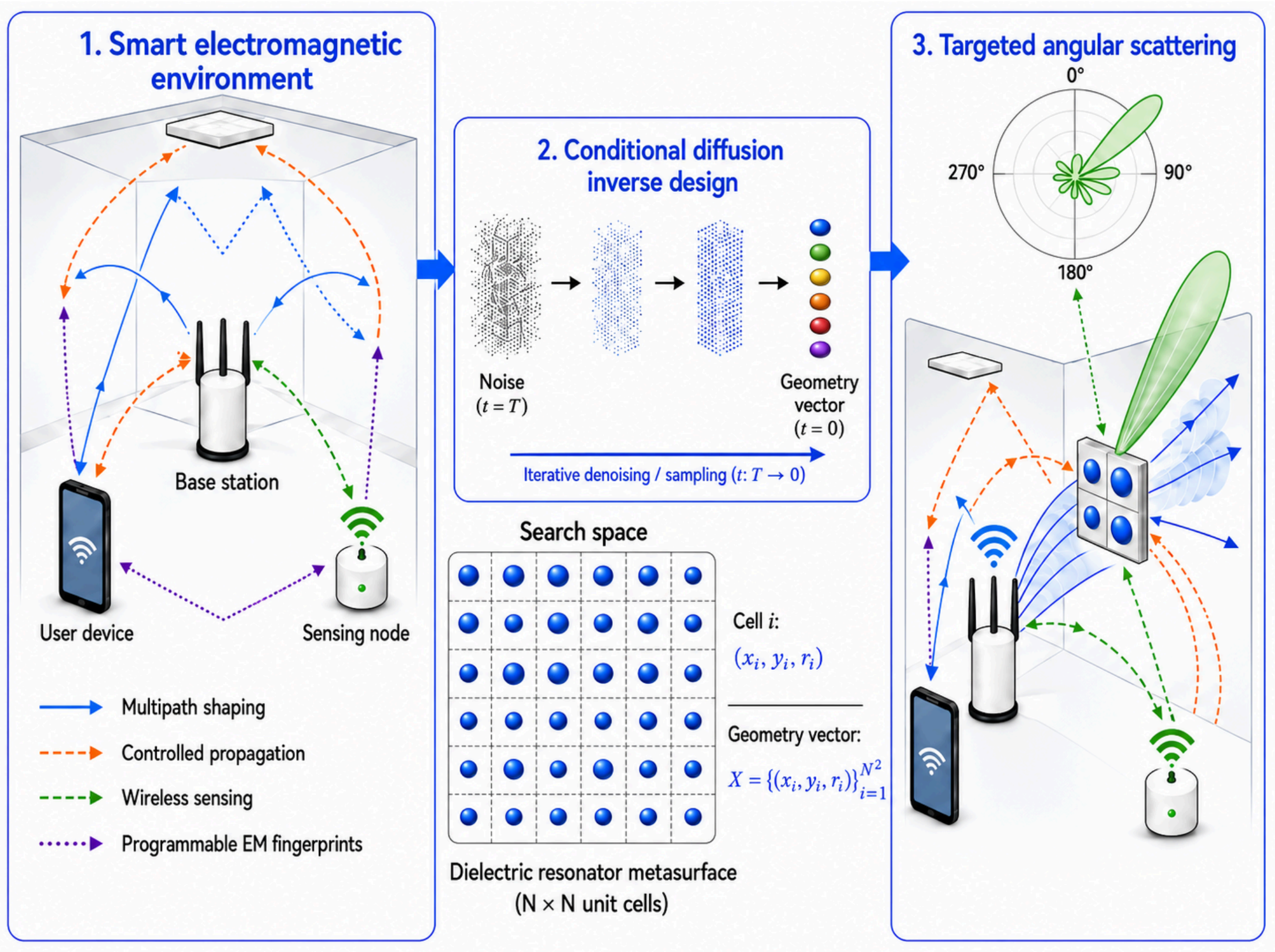

**Figure 1. Diffusion-based inverse design for smart electromagnetic environments. Step 1:** The electromagnetic environment defines the target functionality, including multipath shaping, controlled propagation, wireless sensing, and programmable electromagnetic fingerprints. **Step 2:** The required functionality is encoded as a target angular scattering response and used to condition a diffusion-based inverse-design model. Through iterative denoising, the model generates candidate geometry vectors within an N×N dielectric resonator metasurface search space, where each unit cell contains one spherical resonator described by its in-plane position and radius: $(x_i, y_i, R_i)$. **Step 3:** The generated metasurface is introduced into the environment and verified by forward electromagnetic simulation, yielding a structure that realizes the desired angular scattering response on demand.

The remainder is structured as follows. Angular scattering control is first formulated as an inverse-design problem, followed by a description of the T-matrix dataset generation procedure. The conditional diffusion framework is then introduced and benchmarked against deterministic neural baselines and CMA-ES optimization. Finally, random target-spectrum conditioning is used to evaluate sampling efficiency and generalization.

## Tailoring Angular Scattering as an Inverse Design Problem

Angular control of scattering is usually discussed in terms of beam steering or reflection suppression. However, in many electromagnetic scenarios, the desired response is not limited to forming a single dominant lobe. A scatterer may be required to redistribute energy over several angular sectors, suppress selected directions, enhance side lobes, or create a characteristic angular fingerprint. Such functionality is relevant to smart electromagnetic environments, wireless sensing, identification, calibration objects, and radar-related applications, where the full angular scattering diagram can carry more useful information than one reflected beam. The design objective is therefore formulated as the synthesis of a prescribed angular scattering pattern.

For a given metasurface geometry $x$, the forward electromagnetic model returns the far-field differential scattering cross section. Although the scattering process is fully two-dimensional (it depends on elevation and azimuth spherical coordinates), the design target is reduced here to a one-dimensional polar angular profile. For each geometry, the electromagnetic solver calculates the far-field differential cross section, which is then integrated over the azimuthal coordinate:

$$S_x(\Theta) = \frac{d\sigma_x}{d\Theta} = \int_0^{2\pi} \frac{d\sigma_x}{d\Omega}(\Theta, \varphi) d\varphi. \quad (1)$$

The target response is represented by a vector sampled at a predefined set of polar observation angles:

$$s = [S_{tar}(\theta_1), S_{tar}(\theta_2), ..., S_{tar}(\theta_N)]. \quad (2)$$

Similarly, the response of a candidate structure is written as:

$$f(x) = [S_x(\theta_1), S_x(\theta_2), ..., S_x(\theta_N)]. \quad (3)$$

The inverse design task is to find a physically admissible geometry $x \in X$ whose calculated angular response matches the prescribed vector $s$ . In compact form:

$$x^* = arg\, min_{x \in X} L(s, f(x)), \quad (4)$$

where $X$ denotes the space of allowed metasurface geometries and quantifies the mismatch between the desired and obtained scattering diagrams. This task is non-trivial because the metric compares an entire angular distribution rather than a single scalar quantity. Small deviations near minima, rapid angular variations, and multi-lobe responses can strongly affect the quality of the generated design. In addition, some prescribed patterns may not be achievable within a chosen parameterization, while mathematically suitable candidates may remain detached from practical constraints such as fabrication limits, material availability, losses, and tolerances. Here, the main quantitative error is the mean percentage error (MPE), defined as:

$$MPE(s,f) = \frac{100}{N} | \sum_{i=1}^{N} \frac{s_i - f_i}{s_i} |, \quad (5)$$

where $s_i = S_{target}(\theta_i)$, $f_i = S_x(\theta_i)$. The same metric is used to evaluate generated metasurfaces and compare the diffusion approach with deterministic neural baselines and direct optimization. This formulation makes the design target more general than conventional beam shaping, since the optimization is driven by the complete angular scattering signature.

## Methods

### Forward T-Matrix Model and Dataset Generation

The conditional diffusion model requires paired examples linking a metasurface geometry to its electromagnetic response. In the present work, these pairs are generated synthetically rather than collected experimentally. This choice is deliberate: the forward problem can be solved accurately for the selected class of arrays, allowing a large and internally consistent training set to be produced before the inverse model is trained.

The chosen structures consist of dielectric spherical resonators arranged on a planar (NxN) lattice, as schematically shown in Figure 1. Each unit cell contains one dielectric spherical resonator. The sphere is parameterized by its in-plane center coordinates ($x_i$,$y_i$) and radius ($R_i$), which vary continuously within geometrical bounds that keep it inside the assigned cell and prevent overlap with neighboring elements. The refractive index is fixed at (n=2), corresponding to a non-magnetic lossless dielectric. This value represents a generic moderate-index contrast, relevant to high-index dielectrics in photonics and polymer-based materials at microwave frequencies [23]. All geometrical parameters are normalized to the wavelength. Owing to the scale invariance of Maxwell's equations in dispersionless media, the same formulation can be transferred across frequency ranges, provided that material dispersion, losses, and fabrication constraints are properly accounted for.

The forward solver was implemented using the T-matrix formalism in SMUTHI [24]. This choice is natural for the present parameterization, because each meta-atom is a homogeneous dielectric sphere whose electromagnetic response can be represented by a single-particle scattering operator in a vector-spherical-harmonic basis [25]. The array response is then obtained by coupling these operators through translation matrices, allowing recurrent multiple scattering between all resonators to be included. Compared with mesh-based full-wave simulations, this formulation is computationally efficient for generating large datasets, while still retaining the electromagnetic interactions needed to evaluate the angular scattering pattern. The solver therefore provides a suitable forward model for both training-data generation and independent verification of the structures produced by the inverse-design algorithm.

The dataset used to train the diffusion network was constructed as follows. A small 2×2 array with 4 occupied cells was taken as a benchmark system. Thus, each structure is represented by a 12-dimensional geometry vector. The square design region had a side length of 10λ, and all geometrical parameters were expressed in wavelength-normalized units. For every generated geometry, the forward T-matrix solver calculated the azimuthally integrated far-field response, which was then sampled at 10 polar observation angles: (10°,20°,40°, 60°,70°,80°,100°,120°,140°,160°). The incident field is a TE-polarized plane wave. In total, 11,000 randomly generated structures were calculated, producing paired geometry-response examples for training, validation, and benchmarking.

**Conditional Diffusion Framework for Inverse Scattering Design**

To address the one-to-many nature of inverse geometry design, we propose a conditional diffusion model that learns to generate metasurface geometries from target scattering profiles. As illustrated in Figure 2, the model operates through two complementary phases and follows the Denoising Diffusion Probabilistic Model (DDPM) framework [26]: here, each metasurface geometry is represented by a one-dimensional vector ($x_0$), while the conditioning input is the desired differential scattering cross-section spectrum sampled at selected polar angles. During training, Gaussian noise is progressively added to $x_0$ the forward diffusion process, producing a noisy geometry vector ($x_t$):

$$x_t = \sqrt{\bar{\alpha}_t} \cdot x_0 + \sqrt{1 - \bar{\alpha}_t} \cdot \epsilon;\ \epsilon \sim N(0, I). \tag{6}$$

In the forward diffusion process, the amount of added Gaussian noise was controlled by a cosine schedule, which provides a smoother SNR decay than a linear schedule and improves the stability of DDPM training [27]:

$$f(\tau) = cos^2\left(\frac{\tau+s}{1+s} \cdot \frac{\pi}{2}\right);\ \bar{\alpha}_t = \frac{f(\frac{t}{T})}{f(0)}. \tag{7}$$

The neural network receives the noisy geometry vector ($x_t$), the diffusion step ($t$), and the target spectrum condition. It learns to predict the noise component ($\epsilon_p(t)$) that was added during the forward process at each step. This prediction is used to compute the mean of the reverse Gaussian distribution used for denoising:

$$\mu(x_t, t) = \frac{1}{\sqrt{\alpha_t}}\left(x_t - \frac{\beta_t}{\sqrt{1-\bar{\alpha}_t}}\epsilon_p(t)\right);\ \beta_t = 1 - \alpha_t;\ \alpha_t = \frac{\bar{\alpha}_t}{\bar{\alpha}_{t-1}}. \tag{8}$$

There, $\beta_t$ denotes the forward noise variance. Using $\mu(x_t, t)$, the denoising process is performed, Eq. 9.

$$x_{t-1} = \mu(x_t, t) + \sqrt{\frac{1-\bar{\alpha}_{t-1}}{1-\bar{\alpha}_t}\beta_t} \cdot z;\ z \sim N(0, I), \tag{9}$$

The meaning of the final denoised vector depends on the stage of operation. During training, the clean geometry vector ($x_0$) is known, and the network learns to recover it from a noisy realization by predicting the injected Gaussian noise. During inference, no initial geometry is provided; instead, the reverse process starts from a random Gaussian vector ($x_T \sim N(0, I)$). After the T denoising steps, the model outputs a generated geometry vector ($x_0^*$) conditioned on the prescribed scattering spectrum.

The denoising network is based on a 1D U-Net architecture [28], chosen to operate directly on the geometry-vector representation. The architecture upgrade was performed in the conditioning field. To incorporate the target scattering spectrum into the generation process, FiLM modulation was implemented in the architecture [29]. In this mechanism, the desired spectrum vector is transformed into affine modulation parameters (γ, β) generated by two-layer networks (different for each ResNet Block in the U-Net architecture, 19 in total, producing different γ & β for different channels), processing the scattering conditions. These parameters adjust the internal feature maps of the U-Net during denoising, Eq. 10. The network therefore learns not only the distribution of valid geometry vectors, but also how the denoising trajectory should change for different requested scattering signatures. The advantage of FiLM conditioning over, for example, concatenation of a vector of desired scattering values and a vector of geometric features of a metasurface is the ability to add information about the spectrum to any number of hidden states of a neural network. In this way, information about scattering remains undiminished in the depth of the network, which makes it possible to train deeper and more complex architectures more efficiently.

$$y_{modulated} = y \cdot (1 + \gamma) + \beta;\ y - hidden\ state, \tag{10}$$

The network was trained on the simulated dataset described above using 1000 diffusion steps, a batch size of 16, a learning rate of ($4 \cdot 10^{-6}$), and 116 training epochs. This architecture enables repeated sampling for the same target spectrum, producing multiple candidate geometries that can be ranked by their forward-calculated scattering error. The entire learning process took 6 hours on a virtual machine with access to a 2.20 GHz dual-threaded Intel Xeon CPU E5-2673 v4 processor.

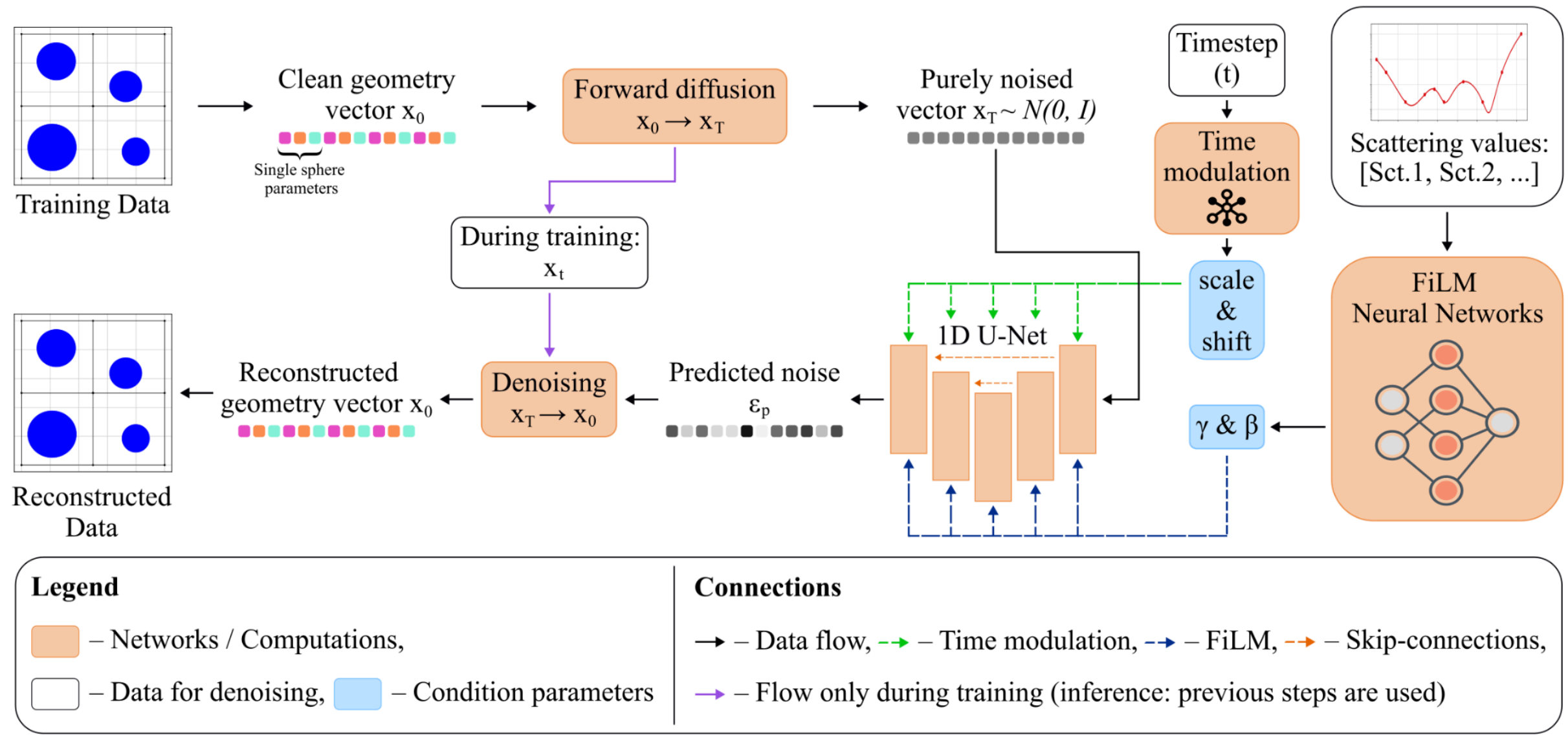

**Figure 2. Conditional Denoising Diffusion Probabilistic Model (DDPM) workflow for metasurface geometry generation.** A dielectric resonator metasurface is first encoded as a

one-dimensional geometry vector. During training, Gaussian noise is added to this vector through the forward diffusion process. A one-dimensional U-Net receives the noisy geometry, the diffusion step, and the desired differential scattering cross-section (DSCS) spectrum, which is introduced through Feature-wise Linear Modulation (FiLM) and Time modulation to know the number of denoising steps. The network predicts the added noise, enabling reconstruction of a clean geometry vector. The reconstructed vector is then converted back into a candidate metasurface geometry.

## Results

### Diffusion Model Performance and Baseline Comparison

Figure 3A shows the evolution of the MPE statistics during training. To test generalization beyond the training set, the model was evaluated on a target scattering spectrum that was not included in the training data. For each training milestone, the conditional generation process was repeated ten times to calculate the mean, median, and standard deviation of the resulting MPE values. The decrease in these statistics with the number of epochs indicates that the model gradually learns to generate geometry vectors whose forward-calculated scattering responses approach the conditioning spectra. This behavior shows that the model does not simply pick geometries randomly from the search space, but learns a reproducible relation between the target scattering response and admissible metasurface geometries. The remaining fluctuations are expected because diffusion sampling is stochastic and because the inverse task requires matching the scattering response at multiple polar angles, rather than optimizing a single scalar quantity.

After training, the diffusion model was used to sample 40 candidate metasurfaces conditioned on the target DSCS profile of the unseen structure. The best-performing generated sample, shown in Figure 3B, achieved an MPE of 1.39%, indicating accurate reproduction of the target scattering response at the specified polar angles, including both low-scattering regions and the rapidly increasing response at larger angles. The MPE distribution across all 40 generated samples is shown in Figure 3C. The median error of 18.91%, together with a relatively compact interquartile range, indicates that the model generates physically relevant candidates rather than isolated successful samples. The spread of the distribution reflects the stochastic nature of diffusion-based generation: individual samples differ in quality, but the generated candidate pool contains several structures that can be ranked by forward electromagnetic verification. This workflow is important for practical inverse design, where the neural model proposes candidate geometries and the final selection is performed using the physical solver.

### Comparison with Deterministic Neural Baselines

The diffusion model was further benchmarked against two deterministic neural baselines: a multilayer perceptron (MLP) and a one-dimensional convolutional neural network (1D-CNN), as described in Appendix A1. Figure 3D compares the MPE distributions obtained for 100 out-of-distribution target spectra. The deterministic models produce broader error distributions and more high-error cases, reflecting the difficulty of mapping a one-to-many inverse problem onto a single predicted geometry. In contrast, the diffusion model yields a lower and more compact error distribution, indicating that conditional generative sampling is better suited for this inverse scattering task.

Overall, Figure 3 demonstrates three key points. First, the diffusion model converges during training toward a lower forward-verified reconstruction error. Second, high-quality geometries can be generated for target spectra that were not directly present in the training set. Third, the probabilistic generation strategy provides a clear advantage over deterministic neural predictors when the objective is to synthesize full angular scattering patterns.

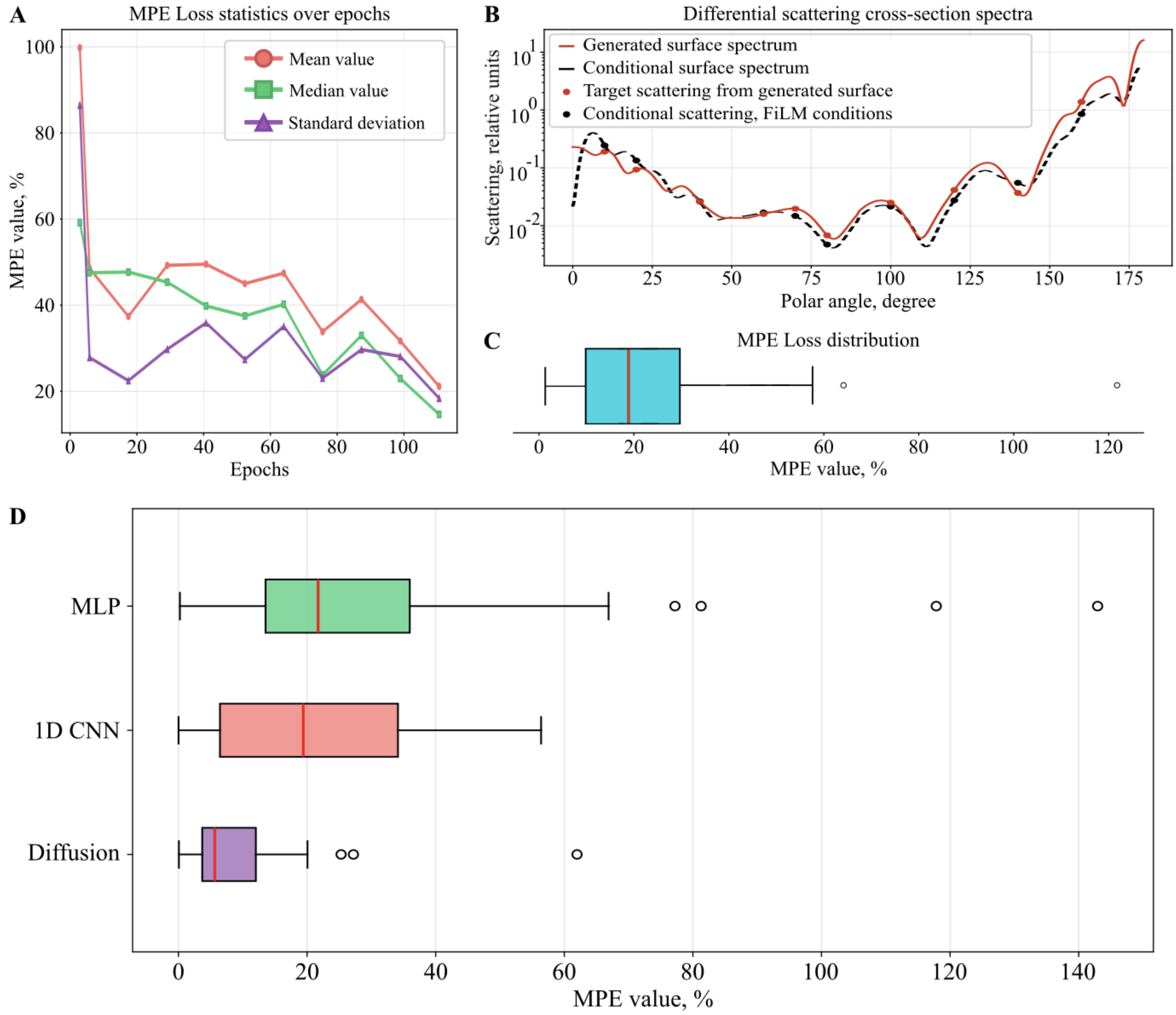

**Figure 3. Performance of the conditional diffusion framework for inverse scattering design.** **(A)** Evolution of the mean percentage error (MPE) during training, showing the mean, median, and standard deviation over generated samples. **(B)** Representative target-spectrum reconstruction, comparing the prescribed differential scattering cross-section (DSCS) profile with the forward-calculated response of the best generated metasurface. **(C)** MPE distribution for 40 independently generated candidates conditioned on the same out-of-distribution target spectrum. **(D)** Benchmark against deterministic multilayer perceptron (MLP) and one-dimensional convolutional neural network (1D-CNN) baselines over 100 out-of-distribution target spectra.

**Comparison with evolutionary optimization**

The performance of the diffusion model was also compared with the evolutionary optimization algorithm CMA-ES, Appendix A2. The evolutionary stochastic optimization algorithm was chosen for comparison as one of the most stable optimization methods with good convergence in complex and nonlinear search spaces, which, however, does not belong to the class of deep learning algorithms. CMA-ES has proven itself well in various metasurface optimization problems in electromagnetism [11], [30]. The scheme of its optimization process is detailed in Appendix A2. CMA-ES starts by randomly generating an initial population of array geometries. Each candidate is evaluated by calculating the MPE between its forward-calculated scattering response and the target spectrum. The candidates are then ranked from lowest to highest MPE, and the best-performing half of the population is selected. This selected subset is used to update the mean and covariance matrix of the sampling distribution, guiding the next generation toward regions of the design space with lower error. Using the updated sampling distribution, CMA-ES generates a new set of candidate geometries. These candidates replace the discarded lower-performing geometries, and the optimization loop is repeated: objective-function evaluation → ranking → selection → distribution update. In this work, the loop was terminated after a preset number of iterations.

The main CMA-ES hyperparameters are the population size, the initial step size (σ), and the number of iterations. The values used in this study are summarized in Table 1. In addition to these parameters, the initial mean of the sampling distribution is important because it determines the region of the design space explored at the beginning of the optimization. A suitable initialization can accelerate convergence, whereas a poor initialization may increase the required optimization time and lead to lower-quality solutions. The population size λ was selected based on the commonly used CMA-ES recommendation as $\lambda = 4 + 3\lfloor log(d) \rfloor$ [31], where $d$ the dimensions of the problem are (here $d = 12$). This choice corresponds to a population size of 12. Convergence usually requires O(100d) to O(1000d) iterations [32]. In this study, two population sizes were tested: the recommended value of 12 and an enlarged population of 60.

| Hyperparameter | Value |
|---|---|
| Population size (λ) | 12 / 60 |
| Value of $\sigma$ | 0.05 |
| Number of iterations | 1,200 |

**Table 1.** Hyperparameters of CMA-ES optimization.

The population size affects both optimization quality and runtime. Larger populations explore a broader region of the design space at each iteration, but require more forward-scattering calculations. The results of the numerical experiments are summarized in Table 2.

| Algorithm / Model | CMA-ES, $\lambda = 12$ | CMA-ES, $\lambda = 60$ | Diffusion |
|---|---|---|---|

| Best solution (MPE) | 9.4% | 4.1% | 1.4% |
|---|---|---|---|
| Runtime | ≈ 2 hours | ≈ 10 hours | ≈ 1 minutes |

**Table 2.** Performance comparison of CMA-ES and the Diffusion model

The diffusion model outperforms CMA-ES both in solution quality and generation time. After a one-time training stage of approximately 6 hours, the trained diffusion model generates candidate metasurfaces for a new target spectrum in less than one minute. In contrast, CMA-ES must be restarted for each new target spectrum and requires repeated forward-scattering evaluations throughout the optimization process, leading to runtimes of several hours. The difference is especially important because the performance of CMA-ES depends on the initial population. Different random initializations can lead to different convergence trajectories and final errors, so several independent restarts are often required to obtain a reliable solution. The diffusion model avoids this repeated optimization cost: once trained, it can be reused for different target spectra and can rapidly provide a pool of candidate geometries for subsequent forward electromagnetic verification.

**Random Spectrum Conditioning and Sampling Efficiency**

To test whether the trained models can respond to scattering requirements beyond spectra directly drawn from the dataset, an additional random-spectrum conditioning study was performed. In this test, target angular responses were generated by varying two parameters of a random spectral construction: the number of terms in the series and the maximal random frequency, Eq. 11. The random target spectra were constructed as weighted sums of angular components with random frequencies and phases. This choice is motivated by the modal nature of electromagnetic scattering, where far-field angular responses can be represented as superpositions of angular modes or multipole contributions. The amplitude weighting (1/k) was introduced as a simple regularization envelope: lower-order components define the dominant scattering lobes, whereas higher-order components contribute weaker fine-scale angular variations. This prevents the synthetic targets from being dominated by rapid oscillations while still providing diverse conditioning spectra for testing the generative model:

$$S(\theta) = \sum_{k=1}^{K} \frac{1}{k}|sin(\omega_k\theta + \varphi_k)|;\ \theta \in [0; \pi];\ \omega_k \sim U(1; \omega_{max});\ \varphi_k \sim U(0; 2\pi), \tag{11}$$

where $U$ denotes a uniform distribution. To recap, the amplitude of the (k)-th angular component was weighted by (1/k), rather than chosen randomly, to suppress excessive high-order oscillations. This weighting is consistent with multimode electromagnetic scattering, where lower-order modes typically define the dominant angular response and higher-order modes add weaker fine-scale variations.

Figure 4A compares the absolute MPE obtained by the multilayer perceptron (MLP), the one-dimensional convolutional neural network (1D-CNN), and the conditional diffusion model for the same set of random target spectra. The horizontal axis shows the number of angular components used to generate the target spectrum, while the vertical axis shows the maximum angular frequency allowed in the random spectrum. Moving to the right increases the number of components, and moving upward increases the angular complexity of the target profile. The color scale represents the absolute MPE, with lower values corresponding to better agreement between the generated and target scattering spectra. The deterministic neural networks show relatively large errors over much of the grid, with several cases exceeding 40–60% MPE. By contrast, the diffusion model maintains substantially lower errors across the tested conditions, including cases with increased spectral complexity. This behavior indicates that the diffusion generator does not merely interpolate a fixed geometry-response map, but can use the conditioning vector to search a broader family of candidate metasurface geometries.

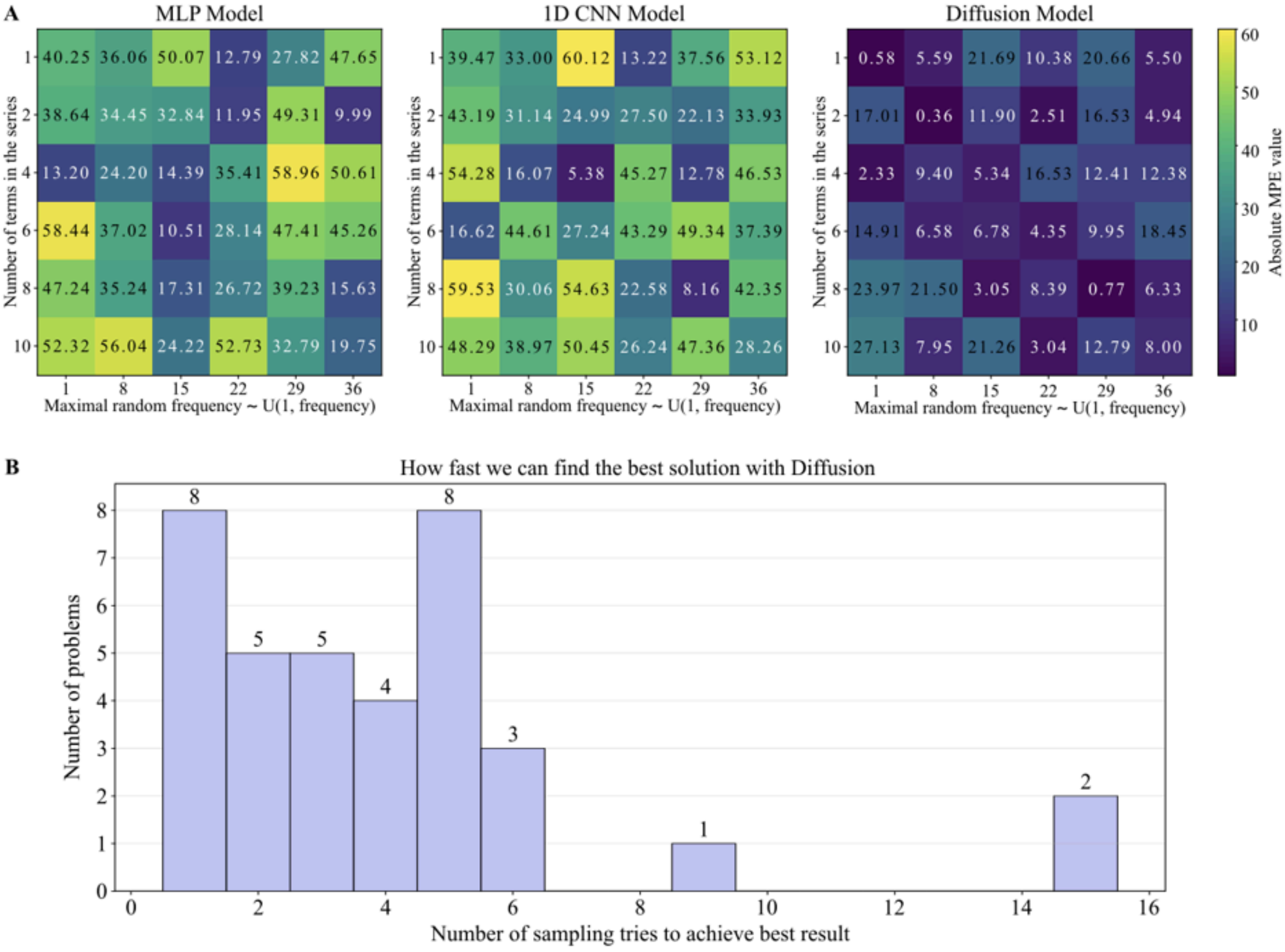

**Figure 4. Random angular scattering targets. (A)** Absolute mean percentage error (MPE) obtained by the multilayer perceptron (MLP), one-dimensional convolutional neural network (1D-CNN), and diffusion model for randomly generated target spectra with different numbers of series terms and maximal random frequencies. **(B)** The number of diffusion sampling attempts required to obtain a generated geometry that outperforms both deterministic neural baselines.

The sampling behavior of the diffusion model is further quantified in Figure 4B. The horizontal axis shows the number of diffusion sampling attempts required to obtain a generated geometry that outperforms the deterministic neural baselines. The vertical axis shows the number of target spectra for which this occurred after the corresponding number of attempts. For each random target spectrum, the model was sampled repeatedly until a generated geometry outperformed both deterministic baselines. In most cases, only a small number of sampling attempts were required, while a few harder conditions required additional trials. This result is important for inverse scattering design: the diffusion model is not limited to a single output, and performance can be improved by generating a candidate pool followed by forward electromagnetic verification. The computational cost of repeated sampling remains low compared with direct iterative optimization, making this workflow suitable for practical screening of multiple metasurface designs. As an intermediate conclusion, this study demonstrates that the conditional diffusion model is more robust to random target scattering conditions than the deterministic baselines. Across the tested grid of random spectra, the MLP and 1D-CNN frequently produce high MPE values, while the diffusion model keeps the error substantially lower for most conditions. This indicates that generative sampling is better suited to the one-to-many nature of the inverse scattering task. The sampling-efficiency analysis further shows that the best diffusion result is usually found after only a few trials, meaning that repeated generation followed by forward verification can improve design quality without resorting to expensive direct optimization. Overall, these results support the use of diffusion models as practical candidate generators for complex metasurface inverse design.

## Conclusion

A conditional diffusion framework was developed for the inverse design of dielectric resonator metasurfaces with prescribed angular scattering responses. The inverse-design task was formulated as the generation of physically admissible geometry vectors whose forward-calculated differential scattering cross-section spectra match a target angular profile. Paired geometry–response data were generated using a T-matrix forward solver, allowing the diffusion model to learn a conditional distribution of metasurface geometries rather than a single deterministic mapping. The results demonstrate that diffusion-based generation is well suited to the one-to-many nature of electromagnetic inverse design. For an unseen target spectrum, the trained model generated a candidate metasurface with an MPE of 1.39% after forward electromagnetic verification. Compared with deterministic MLP and 1D-CNN baselines, the diffusion model produced lower and more compact error distributions for out-of-distribution targets. It also outperformed CMA-ES optimization in both reconstruction quality and runtime after training, reducing the generation of candidate structures from several hours of iterative optimization to approximately one minute of inference. The random-spectrum conditioning study further showed that the model can respond to target angular profiles beyond those directly sampled from the training set. The use of repeated stochastic sampling allowed multiple candidate geometries to be generated for the same prescribed spectrum, with final selection performed by the physical solver. This workflow is important for practical metasurface synthesis, where non-unique solutions, fabrication constraints, and forward-verification requirements make single-output deterministic prediction insufficient.

As an outlook, the main potential of the proposed approach lies in complex electromagnetic environments where direct optimization becomes prohibitively expensive. In realistic environments, each candidate metasurface must be evaluated together with surrounding objects, sources, receivers, and boundary conditions. Repeating such full-wave calculations inside an evolutionary or topology-optimization loop can quickly become impractical. A trained diffusion model can instead serve as a fast candidate generator, producing physically plausible designs that are then screened by a limited number of high-fidelity simulations. This makes the approach attractive for system-level metasurface optimization, where the direct solver is the computational bottleneck.

## Data Availability

The dataset and source code supporting the findings of this study are available at https://github.com/mikzuker/inverse_design_metasurface_generation.

## Author Contributions

M.T. and K.G. conceptualized the study, designed the parametrization, performed diffusion model training and inference, and analyzed results statistically. D.V. coordinated physics aspects of the project. P.G. supervised the project. All authors contributed to writing the manuscript and approved the final version.

## Supplementary information

### A1. Deterministic neural-network baselines

Two deterministic neural architectures were implemented as additional baselines, Figure S1: a multilayer perceptron (MLP) and a one-dimensional convolutional neural network (1D-CNN). Both networks receive the target scattering vector as input and predict a single metasurface geometry vector. The MLP consists of fully connected layers with normalization, nonlinear activation, and dropout. The 1D-CNN uses convolutional blocks to extract local patterns from the input scattering vector before mapping the encoded representation to the geometry parameters. These models test whether the inverse-design task can be solved by direct deterministic regression from scattering spectra to geometries. Their learnable parameter counts were chosen to be comparable to the diffusion model, so that the comparison reflects the effect of the architecture rather than a large difference in network capacity. The MLP contains approximately 935,000 trainable parameters, while the 1D-CNN contains approximately 943,000. The 1D U-Net contains approximately 880,000 trainable parameters.

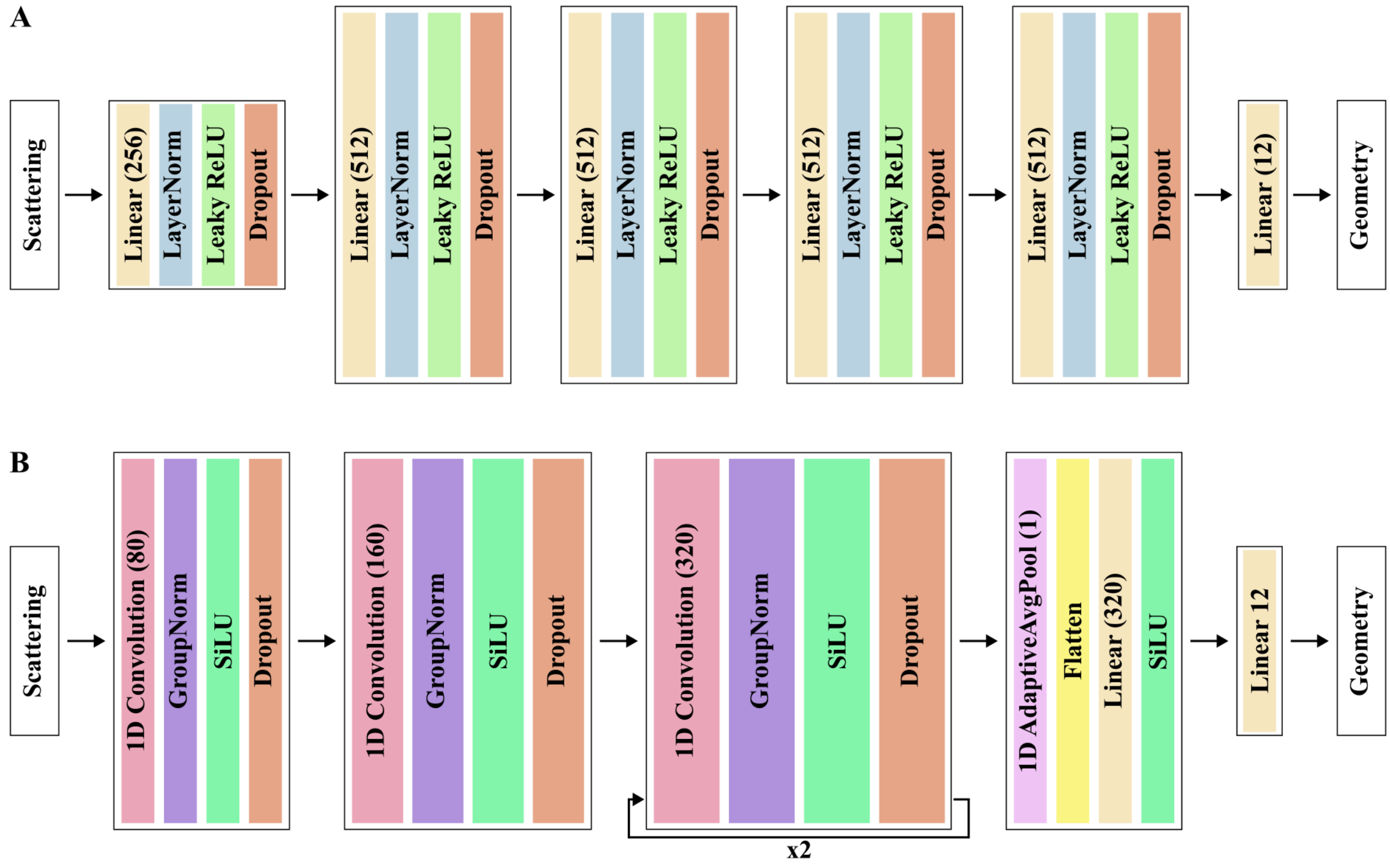

**Figure S1.** Deterministic neural-network baselines. Architectures used for comparison with the conditional diffusion model: (A) A multilayer perceptron (MLP) mapping the scattering vector directly to a metasurface geometry vector; (B) a one-dimensional convolutional neural network (1D-CNN) using convolutional feature extraction followed by a dense output layer. Both architectures produce one deterministic geometry for each prescribed scattering spectrum.

## A2. CMA-ES optimization baseline

The diffusion model was compared with the covariance matrix adaptation evolution strategy (CMA-ES), a derivative-free evolutionary optimization method. CMA-ES was selected as a non-neural baseline because it is suitable for nonlinear search spaces where gradients are unavailable or expensive to compute. In the present inverse-design task, each individual in the population corresponds to a candidate metasurface geometry vector. At every iteration, the forward T-matrix solver evaluates the angular scattering response of each candidate, and the mean percentage error is used as the objective function. The population statistics are then updated by adapting the mean vector and covariance matrix of the sampling distribution, from which the next generation of candidate geometries is drawn. CMA-ES therefore provides a direct optimization reference: unlike the trained diffusion model, it does not learn a reusable inverse map, but solves each target-spectrum problem from scratch through repeated forward simulations. This makes it useful for benchmarking reconstruction quality and runtime against the diffusion-based generator. The algorithm starts from an initialized population of candidate metasurface geometries. Each candidate is evaluated by the forward solver, ranked according to the scattering-error objective, and used to update the population means and covariance matrix. New geometries are then sampled from the updated distribution until the termination criterion is reached, Figure S2.

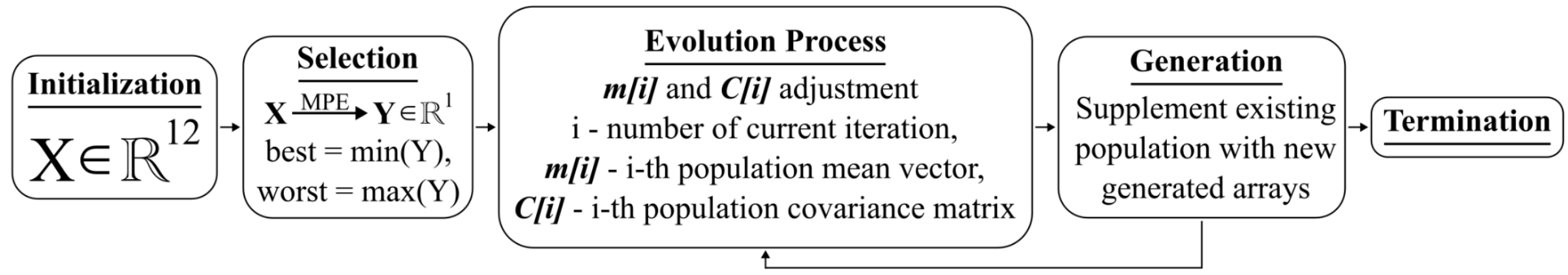

**Figure S2**. Scheme of CMA-ES (Covariance Matrix Adaptation Evolution Strategy) algorithm.